\documentclass[review, compsoc]{IEEEtran}
\usepackage[comma,numbers,square,sort&compress]{natbib}
\usepackage{amsfonts,amssymb,amsthm}
\usepackage{epsfig}
\usepackage{float}
\usepackage{graphicx}
\usepackage{color,hyperref}
\usepackage{multirow}
\usepackage{tipa}
\usepackage{picins}
\usepackage{color}
\usepackage{booktabs}
\usepackage[table]{xcolor}
\usepackage{threeparttable}
\def\BibTeX{{\rm B\kern-.05em{\sc i\kern-.025em b}\kern-.08em
    T\kern-.1667em\lower.7ex\hbox{E}\kern-.125emX}}
\begin{document}

\title{ConMAE: Contour Guided MAE for Unsupervised Vehicle Re-Identification}

\author{Jing Yang, Jianwu Fang, and Hongke Xu
\thanks{This work was supported in part by the National Natural Science Foundation of China under Grants 62036008 and 62273057; in part by the Natural Science Basic Research Plan in Shaanxi Province of China under Grant 2022JM-309.

The authors are with Chang'an University, Xi'an, China, and J. Fang is also a visiting scholar at the NExT++ Research Centre of the School of Computing, National University of Singapore, Singapore.
        {(fangjianwu@chd.edu.cn)}.}%           
  }

\IEEEtitleabstractindextext{%
\begin{abstract}
Vehicle re-identification is a cross-view search task by matching the same target vehicle from different perspectives. It serves an important role in road-vehicle collaboration and intelligent road control. With the large-scale and dynamic road environment, the paradigm of supervised vehicle re-identification shows limited scalability because of the heavy reliance on large-scale annotated datasets. Therefore, the unsupervised vehicle re-identification with stronger cross-scene generalization ability has attracted more attention. Considering that Masked Autoencoder (MAE) has shown excellent performance in self-supervised learning, this work designs a Contour Guided Masked Autoencoder for Unsupervised Vehicle Re-Identification (ConMAE), which is inspired by extracting the informative contour clue to highlight the key regions for cross-view correlation. ConMAE is implemented by preserving the image blocks with contour pixels and randomly masking the blocks with smooth textures. In addition, to improve the quality of pseudo labels of vehicles for unsupervised re-identification, we design a label softening strategy and adaptively update the label with the increase of training steps. We carry out experiments on VeRi-776 and VehicleID datasets, and a significant performance improvement is obtained by the comparison with the state-of-the-art unsupervised vehicle re-identification methods. The code is available on the website of \url{https://github.com/2020132075/ConMAE}
\end{abstract}

\begin{IEEEkeywords}
Masked autoencoder (MAE), Vehicle re-identification, Contour guidance, Unsupervised learning
\vspace{4em}
\end{IEEEkeywords}}

\maketitle

\section{Introduction}

Vehicle re-identification involves a cross-view vehicle image matching problem, which is the key technique of intelligent traffic management and control systems, especially for the cross-localization of important vehicles in the road network. Compared with the well-developed pedestrian re-identification models \cite{c28,c29}, vehicles are more similar in appearance, which causes small inter-class differences. In addition, the different directions of the vehicle make the intra-class difference larger than the one of pedestrians. 

The main task in vehicle re-identification is to learn a feature representation with a small distance between the same vehicle in different views and a large distance between different vehicles \cite{c6}. During the decades, the paradigms of vehicle re-identification can be divided into supervised methods \cite{c3,c4,c31} and unsupervised approaches \cite{c32,c33,c35} according to whether the label of training data is involved or not. The supervised vehicle re-identification relies on the labels of the vehicle images, and approximates the fixed distribution of the training samples \cite{c37,c36}. Apparently, it may be influenced by the out-of-distribution (OOD) samples \cite{fort2021exploring} in the testing phase. On the contrary, unsupervised vehicle re-identification aims to cluster the samples and update their pseudo labels with the training stage \cite{c32,c33}. Because the training is supervised by pseudo labels of vehicle images, the accuracy may be limited compared with supervised vehicle re-identification models. However, the unsupervised mode has better generalization for different occasions and is easy to be implemented with no annotation works. In this work, we will explore unsupervised vehicle re-identification.

\begin{figure}[!t]
\centering
\includegraphics[width=\hsize]{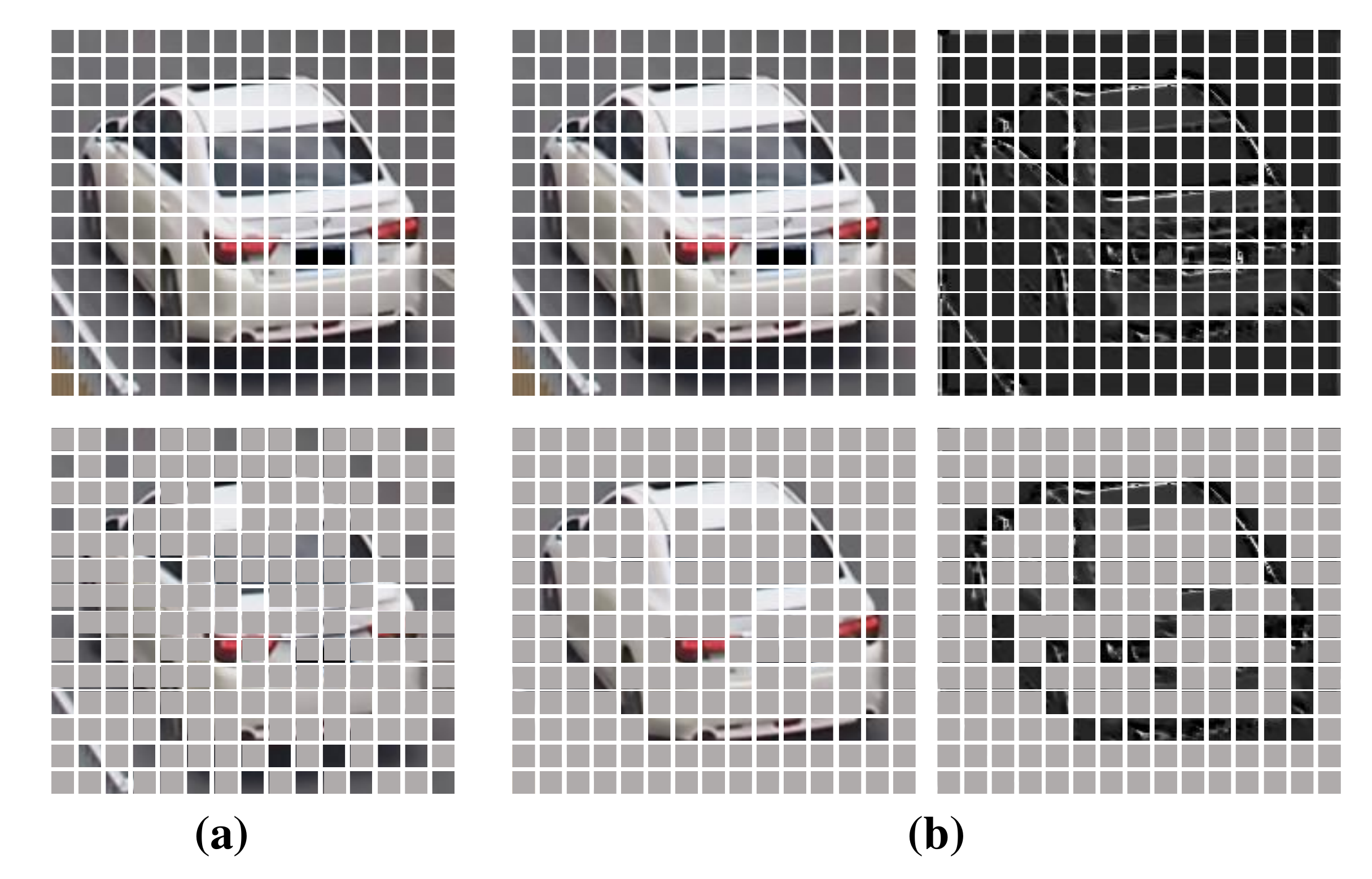}
\caption{\small{The idea of ConMAE, where (a) denotes the random masking strategy on the whole image, and (b) is the masking on the smooth regions while preserving the contour regions.}}
\label{fig1}
\end{figure}
In unsupervised re-identification, the key issues are to learn a good feature representation and improve the quality of pseudo labels. Recently, self-supervised Masked Autoencoder (MAE) \cite{c16} has achieved great success in feature representation and assisted many downstream tasks. Therefore, we take MAE as a backbone model to obtain the feature representation of vehicle images. MAE owns a random masking mechanism for enhancing the learning of the local image patches for the whole image (as shown in Fig. \ref{fig1}(a)). It is fulfilled by encoding the masked image with some unmasked patches preserved, and then decoding the original image using the unmasked patches. Through this process, self-supervised representation learning is achieved. As for vehicle re-identification, many different vehicles have similar textures and appearances for shape, color, and view. Maybe, the difference is reflected by some tiny or contour regions. Therefore, in this work, we aim to enforce the local contour regions in MAE learning. Consequently, we propose a contour-guided MAE (ConMAE) for this purpose. As shown in Fig. \ref{fig1}(b), we can see that we preserve the patches with contour pixels, and the structure of the vehicle is maintained better than the original masking strategy in MAE. ConMAE is fulfilled by contour detection with convolution on original images and then utilizes the distance transform to enhance the intensity of contour pixels. With that, the random masking is taken on the smooth image patches and preserves the contour region in the encoding process of MAE. 

With the feature representation, we introduce a progressive pseudo label updating with the increase of training epochs. Different from other works that use clustered pseudo labels with no link to the previous training epochs, progressive updating can soften the pseudo labels and improve the quality of the pseudo labels. Finally, we construct the feature dictionary by the cluster center in the encoded features of the training samples and fulfill the vehicle re-identification model by calculating and minimizing the soft contrast loss of all training samples. The main \textbf{contributions} of this paper are as follows. (1) A contour-guided Masked Autoencoder (ConMAE) is proposed, which can make the encoding process pay more attention to the contour pixels of vehicle images. It is particularly suitable for the downstream vehicle re-identification task with similar appearance, view, and shape issues. (2) We design a progressive pseudo label softening mechanism to improve the quality of the pseudo label, which eliminates the fluctuation of the hard (without link with previous training epochs) pseudo labels and avoids the re-identification confusion by the pseudo labels.

\section{Related Works}
\subsection{Supervised Methods}
The supervised vehicle re-identification approach has achieved good results in this field. The main formulations are to find the better feature representation in local or global views \cite{c39,c41}, model the relationship among samples \cite{c40,c42}, and enhance the diversity of samples with some data generation frameworks \cite{c43,c44}. For example, the works of \cite{c1,c2} improve the feature representation by obtaining a better local view for a part-based feature representation extraction. Then, the local features are combined with the global features (whole image) to form more good fusion features. The accuracy of the re-identification is improved by using the fused features.

There are also some methods that focus on the relationship modeling between samples and optimize the results of vehicle re-identification by learning the relationship distribution of vehicle images \cite{c7}.
Data enhancement is also a framework to augment the diversity of the samples, so as to cover the data distribution of the vehicle images. For example, Zhu \emph{et al.} \cite{c9} improve the Generative Adversarial Network (GAN) to reconstruct images, and use the enhanced data to train the vehicle re-identification model. These data argumentation methods are helpful for a robust model.

However, training these models requires large-scale manually labeled datasets, which leads to some limitations in vehicle re-identification algorithms in actual traffic scenes. Therefore, researchers begin to pay more attention to unsupervised vehicle re-identification recently. Benefiting from the emergence of various new feature extraction models, unsupervised vehicle re-identification has also made great progress.

\begin{figure*}[!t]
\centering
\includegraphics[width=0.9\hsize]{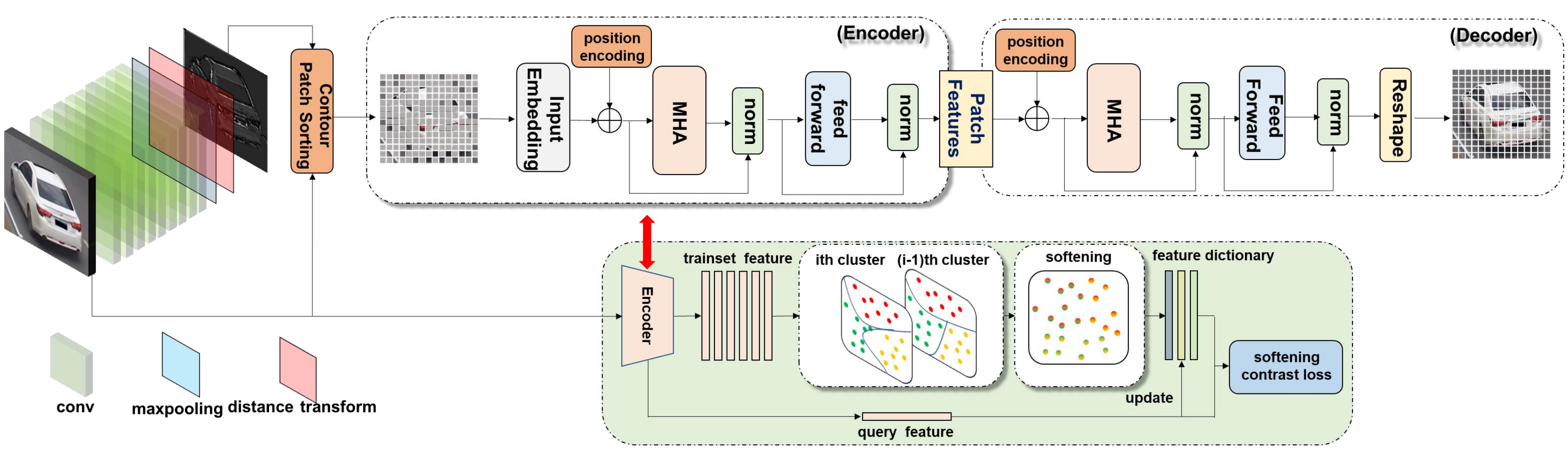}
\caption{Contour Guided MAE for Unsupervised Vehicle Re-Identification.}
\label{fig2}
\end{figure*}
\subsection{Unsupervised Approaches}
Unsupervised learning vehicle re-identification can be divided into domain adaptation and pseudo-label optimization approaches according to whether the training data contains the labeled source domain data. Vehicle re-identification based on domain adaptation usually uses the labeled source domain dataset to pre-train the vehicle re-identification model, and then migrates the pre-trained model to the target domain dataset. It fulfills the learning of the distinguishing features from unlabeled vehicle images. Some works use the generative adversarial network to realize the domain adaptation vehicle re-identification \cite{c11,c12}. These methods learn the domain invariant feature representation by minimizing the feature representation difference between the source domain and the target domain by adversarial training. On the basis of domain adaptation, the accuracy of unsupervised vehicle re-identification has been further improved.

The unsupervised vehicle re-identification using pseudo-label training is now widely used in this field \cite{c14,c15}. This kind of method generally generates pseudo labels for the samples by clustering methods and constructs a feature dictionary according to the pseudo labels. The model is trained by calculating the loss between the feature of the training sample and the feature dictionary. The feature dictionary will be updated during the training, and the accuracy of the pseudo labels will be optimized with continuous iteration. Although unsupervised vehicle re-identification methods have made impressive progress, the accuracy of these models still needs to be improved compared with the supervised vehicle re-identification methods. The accuracy of unsupervised vehicle re-identification using pseudo-label training is closely related to the quality of the pseudo-labels. A good initialization model can generate pseudo labels with high confidence at the beginning, and accelerate the convergence of the model.\\
 
\section{Approach}
 
This paper designs a contour-guided Masked Autoencoder (ConMAE) for unsupervised vehicle re-identification. The purpose of ConMAE is to obtain the encoder that can focus on the key areas of the vehicle image through the process of self-supervised image reconstruction. The encoder is applied to the vehicle re-identification task as a feature extraction model. After unsupervised training, the features of the same vehicle IDs will be gathered together in the feature space, and the features of different vehicle IDs will be separated from each other in the feature space.

As shown in Fig. \ref{fig2}, the proposed method consists of two parts: contour-guided Masked Autoencoder for the feature and unsupervised vehicle recognition as a downstream task. In the self-supervised training phase, the vehicle contour is highlighted under the distance transform method. The random masking of the patch-wise vehicle image is guided by the contour regions. Among them, the patches that are not masked are embedded by the encoder, and the original vehicle image is reconstructed by the decoder. The mean square error loss of the pixels of the reconstructed image and the obscured image patch of the original image is calculated to train the ConMAE. 

In the vehicle re-identification stage, the encoder of ConMAE is used as the feature extraction model to extract the features of all samples of the training set, and then the features are clustered and pseudo labels are assigned. Pseudo-labels will be optimized according to the proposed pseudo-label softening mechanism. The feature dictionary is initialized according to the pseudo label and the clustered feature center of the samples, and the soften cluster contrast loss training model is calculated by using the class center feature stored in the dictionary and the training sample feature. At the same time, the class center feature will also be updated.

\subsection{ConMAE}

ConMAE is a method that preserves the key areas of the image-guided by the contour of the vehicle in the image and reconstructs the image with the unmasked patches with contour pixels. As ConMAE reconstructs the image according to the key areas of the image, its encoder will learn a powerful feature representation during the self-supervised training process, which is more suitable for downstream tasks.

ConMAE includes two parts: the contour guided mask part and the image reconstruction part. In the contour-guided mask part, the smooth regions will be masked first, and the unmasked contour regions will be used for image reconstruction. In the image reconstruction part, the encoder extracts the features of unmasked image patches and decodes the feature vector to the original image.

In the contour-guided mask part, we use the convolution layer, max-pooling layer, and distance transform to mine the contour of vehicles in the image. For the input image, we use $9$ convolution layers with residual connection to mine image features, where the convolution kernel size is $3\times3$ and the step size of $1$ are adopted. Because we only want to extract the edge information in the convolution layer, the input and output of the first eight convolution layers are with three channels, and the output of the $9^{th}$ convolution layer owns a single channel. In addition, we also use the max-pooling layer to highlight the edge information in the convolution maps. On this basis, we use distance transform \cite {c17} to enhance the contour of the vehicle image. Distance transform describes the distance from a non-zero pixel to the nearest zero pixels in an image. The farther the point in the image is from the background pixel, the brighter it is.

We divide the pixels on the obtained feature map of an image into a smooth background set $\omega^b$ (a set of points with zero pixel values) and a target set $\omega^c $(a set of non-zero pixels). As shown in Eq. \ref{eq1}, the purpose of the distance transform is to generate a mapping $D$, which maps each pixel value $p$ of the target set with the minimum distance to the background set $\omega^b$.
\begin{equation}
\label{eq1}
D(p)\ =min(dist(p,q))\ \ \ \ \ p\epsilon\omega^c,q\epsilon\omega^b
\end{equation}
where $dist(p,q)$ generally uses Euclidean distance as:
\begin{equation}
\label{eq2}
disf(p,q)\ =\sqrt{{(p_x-q_x)}^2+{(p_y-q_y)}^2}
\end{equation}
where $(p_x,p_y)$ is the coordinate of the pixel $p$ in $\omega^c$, and $(q_x,q_y)$ is the coordinate of the pixel $q$ in $\omega^b$.

In the specific implementation, the internal points, external points, and isolated points in the image target set are counted first \footnote{The interior points refer to the points whose neighborhoods are non-zero pixels. The isolated points refer to the points whose neighborhoods are all zero pixels, and the other points are denoted as the external points}. For each interior point, the distance between the point and all non-interior points is calculated, and the minimum distance is assigned to the interior point. For the isolated point, the pixel value is kept unchanged. After traversing all pixels in the target set to complete the distance transformation, the vehicle contour in the image after the distance transformation is highlighted.

\begin{figure}[!t]
\centering
\includegraphics[width=\hsize]{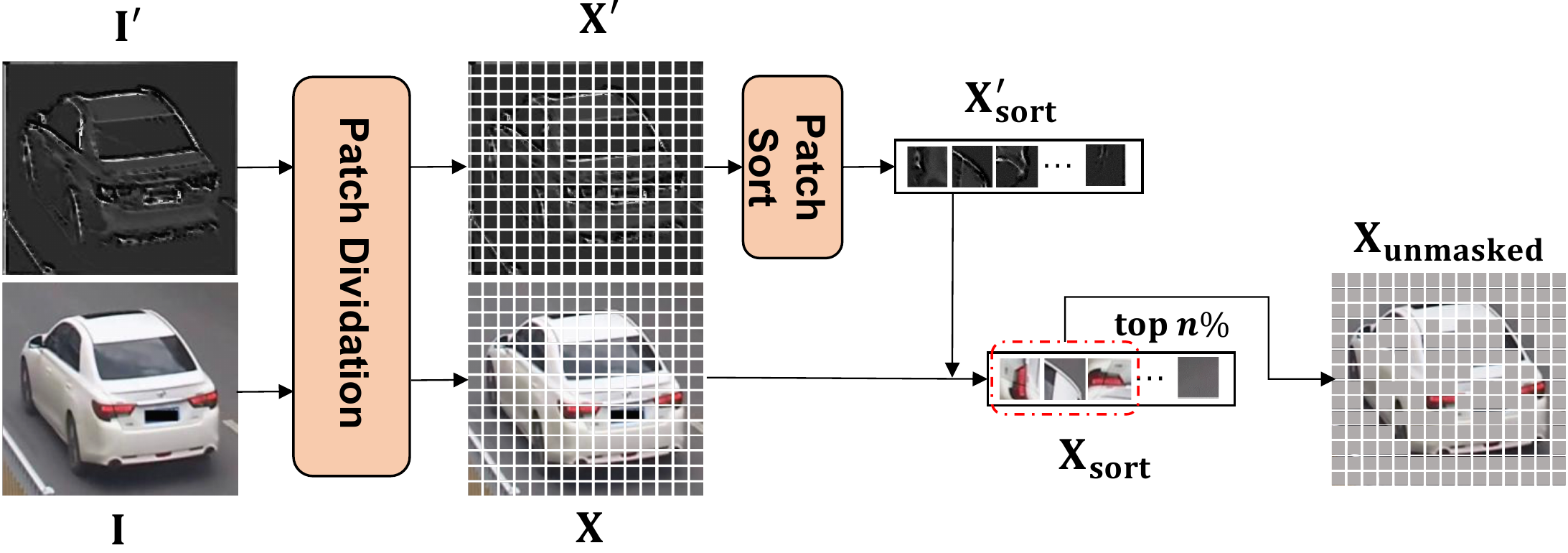}
\caption{\small{Masking operation with contour guided patch sorting.}}
\label{fig3}
\end{figure}
As shown in  Fig. \ref{fig3}, the input original image $I$ and the distance-transformed contour image $I^{\prime}$ are divided into many patches in the same way, which are recorded as the sequence of $X$ and the sequence of $X$ with $K$ patches ($K$=$14\times14$). The contour intensity within each patch can be calculated by the average of the pixel values, and the patches are sorted according to the contour intensity. The sorted patch set is denoted as $X^\prime$. After the sorting procedure for the patches of the contour image, assume we can obtain the indexes of $m$ patches with the top $m$ large contour intensity values in the sorted list $X_{sort}$. It means that these $m$ patches have more contour information than other patches. Except for these $m$ patches, the masking operation of MAE is adopted on the other patches in $X_{sort}$. The $m$ patches then form the unmasked image $X_{masked}$, which is used to contribute to the feature embedding of the vehicle image and reconstruct the original vehicle image.

In the image reconstruction part, we use a similar encoder-decoder structure in MAE \cite{c21}. In the encoder part, each patch in $X_{unmasked}$ will be encoded into a feature by a fully connected network with position encoding. Multi-Head Self-Attention (MHSA) model \cite{DBLP:conf/nips/VaswaniSPUJGKP17} is used to focus on the key information. We train the ConMAE model by calculating the MSE loss between the predicted pixels and the ones in the masked patches of the original image.

\subsection{Unsupervised Vehicle Re-Identification}

In this paper, we design a pseudo-label softening mechanism for vehicle re-identification. We use the encoder of ConMAE as the feature extraction model that embeds the vehicle image into the feature space. After the training, the features belonging to the same vehicle in the feature space will be grouped together, and the features belonging to different vehicles will be separated from each other.

At the beginning of each epoch, we use the encoder to extract the features of all samples in the training set. According to the Jaccard distance \cite{c20} between samples, a clustering algorithm is used to assign pseudo labels to each sample. The Jaccard distance $d\left(f_i,f_j\right)$ between feature $f_i$ and feature $f_j$ can be calculated according to: 
\begin{equation}
\label{eq3}
d\left(f_i,f_j\right)=1-\frac{\sum_{1}^{d}{a_xb_x}}{\sum_{1}^{d}{a_x}^2+\sum_{1}^{d}{b_x}^2-\sum_{1}^{d}{a_xb_x}}
\end{equation}
where $d\left(f_i,f_j\right)$ denotes the Jaccard distance between the features $f_i$ and $f_j$, where $d$ denotes the dimension of the feature, $a_x$ and $b_x$ denote the values in feature $f_i$ and $f_j$ with the index of $x$ ($x\in[1,d]$), respectively.

DBSCAN \cite{c19} here is adopted to cluster the features of vehicle images, which is a density clustering that can adaptively adjust the number of clustering categories, and is widely used in unsupervised learning. It checks the Jaccard distance matrix line by line when assigning hard pseudo labels to samples. 

The hard pseudo labels generated by clustering are difficult to avoid some kinds of deviations caused by hard samples, and the wrong hard pseudo labels have a great negative impact on the model training. In addition, each training step with hard pseudo labels will re-generate the hard pseudo label, which results in the useful information of the previous epoch pseudo label cannot be transferred to the next epoch. Therefore, soft labels have been focused on and can reduce the adverse effects of hard pseudo labels that are incorrectly classified.  

This paper proposes a pseudo-label softening mechanism for unsupervised vehicle re-identification. We broadcast the pseudo labels of the previous epoch to the current epoch by momentum update\cite{c15}, which softens the hard pseudo label generated by the clustering process. The latest pseudo labels always contain some information about the previous pseudo labels. As the training progresses, the pseudo labels will have higher confidence.

We represent the hard pseudo label in the one-hot form generated by the $i^{th}$ image at the $t^{th}$ epoch as $y^{h}_{i_t}$, then the soft pseudo label $y^{s}_{i_t}$ optimized from $y_{i_t}$ at the $t^{th}$ epoch is:
\begin{equation}
\label{eq4}
y^{s}_{i_t}=\lambda\cdot{y^{h}_{i_t}}+\left(1-\lambda\right)\cdot{y^{h}_{i_{t-1}}},	
\end{equation}
where $\lambda$ is a constant and selected as $0.5$ in the experiment.

 \begin{table*}[!t]\small
\centering
\caption{The performance (\%) between ConMAE and some state-of-the-art methods on VeRi-776 dataset and VehicleID dataset.}
\label{tab1}
\begin{tabular}{c|ccc|ccc|ccc|ccc }
\toprule[0.8pt]
\multirow{2}{*}{methods} & \multicolumn{3}{c|}{VeRi-776}                  & \multicolumn{3}{c|}{VehicleID (test size=800)}                                       &
\multicolumn{3}{c|}{VehicleID (test size=800)}                                       &
\multicolumn{3}{c}{VehicleID (test size=800)}    \\ \cline{2-13}
                         & mAP       & Rank-1    & Rank-5   & mAP      & Rank-1    & Rank-5   & mAP       & Rank-1    & Rank-5    & mAP       & Rank-1    & Rank-5 \\ \hline
PUL{\cite{c24}}              & 17.1         & 55.2         & 66.3         & 43.9         & 40.0          & 56.0      & 37.7         & 33.8          & 49.7          & 34.7         & 30.9          & 47.2      \\ 
DAVR{\cite{peng2020cross}}              & 26.4         & 62.2         & 73.7         & 54.0         & 49.5          & \textbf{68.7}      & 49.7         & 45.2          & \textbf{64.0}          & 45.2         & 40.7          & 59.0      \\ 
SPCL{\cite{ge2020self}}     & 36.9 & 79.9          & 86.8             & -          & -           & -             & -             & -              & -              & -             & -              & -      \\ 
PAL{\cite{c25}}              & 42.0         & 68.2         & 79.9         & 53.5         & \textbf{50.3}          & 64.9          & 48.1         & 44.3          & 60.9          & 45.1         & 41.1          & \textbf{59.1} \\ 
\hline
ConMAE                   & \textbf{44.2}          & \textbf{82.7} & \textbf{87.9}          & \textbf{56.3} & 49.7           & 63.5          & \textbf{54.2}             & \textbf{47.5}             & 62.0              & \textbf{50.1}              & \textbf{43.2}             & 57.9              \\ 
\hline
\end{tabular}
\end{table*}
As for the feature dictionary, it is initially constructed by the hard pseudo labels, and the average feature of all sample features with the same hard pseudo labels is denoted as the center feature of the category. The center feature stored in the feature dictionary $F_c$ can be calculated by:
\begin{equation}
\label{eq5}
F_c=\frac{1}{m}\sum_{i\epsilon c}f_i,
 \end{equation}
where $c$ is the category corresponding to the $i^{th}$ image, $m$ is 
number of samples in category $c$, and $f_i$ is the feature of the $i^{th}$ image.

During training, each training sample will be close to the center of the corresponding category and far away from the center of unrelated categories. The center feature will be updated according to the features of training samples by:
\begin{equation}
\label{eq6}
F_c\gets \sigma\cdot F_c+\left(1-\sigma\right)f_i,
 \end{equation}
%where $F_c$ is the center feature stored in the feature dictionary for the category corresponding to the $i^{th}$ image, $f_i$ is the feature of original image $I$, and 
where $\sigma$ is a super-parameter and set as 0.5 in the experiment.

In order to use the soft label training model with higher confidence, we use the softening contrast loss function. We represent the soft pseudo label $y^{s}_{i}$ of the $i^{th}$ image as:
\begin{equation}
\label{eq7}
y^{s}_{i}=\left(p_i^1,p_i^2,p_i^3\ldots p_i^\alpha\right),  	
\end{equation}
where $p_i^\alpha$ is the soft label of the $i^{th}$ image, and $\alpha$ is the number of categories. Then the softening contrast loss $L$ can be described by:
\begin{equation}
\label{eq8}
  L=\sum{-p_i^\alpha\cdot\log{\frac{{\exp{\left(f_i\cdot F_c\right)}}/{\tau}}{\sum_{c=1}^{c=\alpha}{\exp{\left(f_i\cdot F_c\right)}}/{\tau}}}},
\end{equation}
where $\tau$ is the temperature coefficient and is set as $0.05$ in the experiment. 

\section{Experiments}
\subsection{Dataset and Metrics}
The proposed method is tested on the VeRi-776 dataset and VehicleID dataset. 

\textbf{VeRi-776} \cite{c22} is a large-scale vehicle re-identification dataset. It has more than 50,000 pictures of 776 vehicles from 20 cameras. The vehicle pictures are collected from the real world, and the vehicle's identity, camera number, and timestamp information are manually labeled.

\textbf{VehicleID} \cite{c23} has more than 221,763 pictures of 26,267 vehicles. This dataset also includes the type information of some vehicles.

We evaluate the performance of the proposed model by two evaluation indexes: mean Average Precision (\textbf{mAP}) and Cumulative Matching Characteristics (\textbf{Rank-\emph{n}}).

The mAP metric is an indicator to measure the quality of a multiple-category model in all categories, which prefers a larger value for better performance.

Rank-$n$ reflects the probability that the $n$ images with the highest confidence in the retrieval results have correct labels. Rank-$n$ also prefers the large value for the re-identification performance.

\subsection{Implementation Details}
We use the ViT-B/16 pre-trained on ImageNet-1K (IN1K) \cite{c21} as the backbone model. Then we carry out self-supervised pre-training on the VeRi-776 dataset and VehicleID dataset, respectively, and then apply the pre-trained model to the vehicle re-identification task for unsupervised training.

\subsubsection{Experimental Setting on ConMAE training} In the pre-training stage, we use the Adam optimizer whose weight decay is 5e-2 and set the initial learning rate as 2.5e-4. The batch size we set is 32, and the maximum number of the epoch is set as 200.

\subsubsection{Experimental Setting on Unsupervised Vehicle Re-ID} In the unsupervised training stage, we also use the Adam optimizer whose weight decay is 5e-4, and set the initial learning rate of 1.5e-5. The batch size we set is 32, and the maximum number of epochs is 50.

\subsection{Comparison Methods}
We evaluate the experiments on two datasets and compare the results with the state-of-the-art unsupervised vehicle re-identification methods. The experimental results are shown in Table~\ref{tab1}.
 
The effectiveness of ConMAE is proved by the comparison with other methods. Among them, PUL \cite{c25} is a classical unsupervised vehicle re-identification method, which uses progressive unsupervised learning to migrate the model to the target domain. It belongs to the early work in the field of re-identification. On the basis of these classical methods, the research on unsupervised re-identification is more and more in-depth. DAVR \cite{peng2020cross} uses the dual branch adversarial network to make the source domain image learn the style of the target domain image, and train the unsupervised vehicle re-identification model. SPCL \cite{ge2020self} proposes a self-paced competitive learning framework with hybrid memory to gradually create clusters with higher confidence, dynamically generate supervision signals, and improve the model performance. PAL \cite{c25} infers the pseudo labels of the target domain samples according to the source domain distribution, and further improves the model performance through unsupervised training in the target domain.

ConMAE improves the feature representation ability of the feature extraction model through self-supervised pre-training to improve the accuracy of the pseudo label. In addition, the accuracy of the model is improved by softening the pseudo label. The experimental results show that ConMAE has good performance on both the VeRi-776 dataset and the VehicleID dataset. Especially on the VeRi-776 dataset, since there are a large number of samples in each category of the dataset, the model can better focus on intra-class differences when unsupervised training. Therefore, the model performs well on Rank-$n$ metrics. In contrast, the VehicleID dataset has richer sample types, but fewer samples for each category. Therefore, the model can better excavate the differences between classes in unsupervised training, and perform better in the mAP metric.

\subsection{Ablation Studies}

\subsubsection{Compared with ResNet Backbone}
In order to study the impact of different feature extraction models on unsupervised vehicle re-identification tasks, we carry out experiments on different backbone models, and the results are shown in Table~\ref{tab2}.

It can be seen from the experimental results that the model using ConMAE as the backbone is superior to the model using ResNet-50 as the backbone in terms of mAP but inferior to the model using ResNet-50 as the backbone in terms of rank-n. Compared with ResNet-50, the ConMAE encoder, which has been fully self-supervised and pre-trained is better at mining the overall distribution because the pseudo labels generated at the beginning of training have higher accuracy. However, ResNet is more suitable for mining the details of images because of the convolution, so it is better for rank-$n$.
\begin{table}[!t]\small
\centering
\caption{Performance comparison with ResNet backbone on VeRi-776 dataset.}
\label{tab2}
\begin{tabular}{c|cccc}
\toprule[0.8pt]
method    & mAP       & Rank-1    & Rank-5    & Rank-10   \\ \hline
ResNet-50 & 42.5          & \textbf{87.7} & \textbf{91.4} & \textbf{93.1} \\
ConMAE    & \textbf{44.2} & 82.7          & 87.9          & 90.6          \\ \hline
\end{tabular}
\end{table}

\subsubsection{Effects of Contour Guided Mask and Random Mask}
In order to study the influence of the contour guide mask on the model performance, we carry out experiments on the MAE and ConMAE, and the results are shown in Table~\ref{tab3}.

\begin{figure}[htpb]
\centering
\includegraphics[width=0.9\hsize]{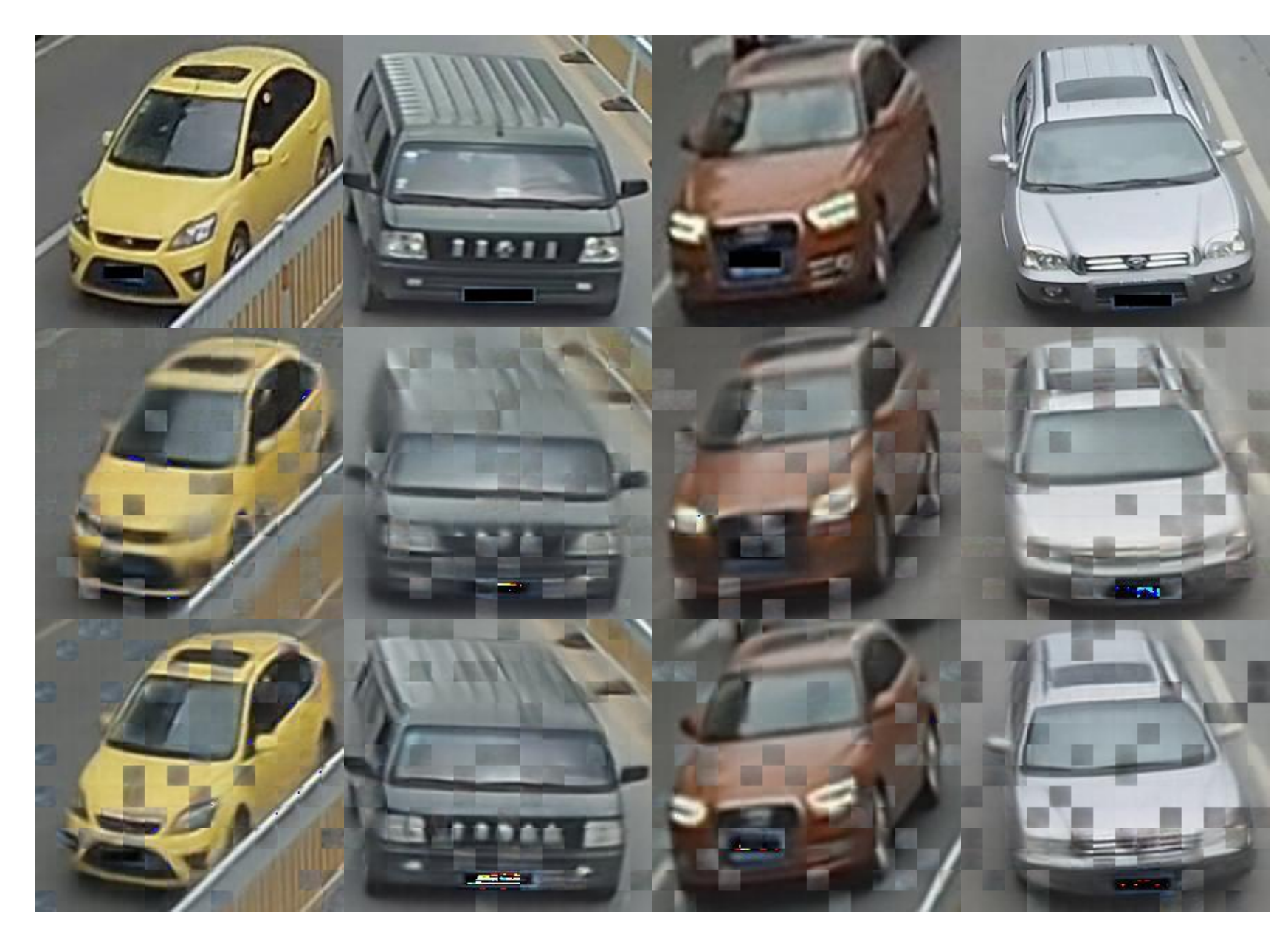}
\caption{\small{From the top to the bottom rows, the images are the original image, the reconstructed images by MAE, and the reconstructed images by ConMAE, where the gray patches are the unmarked patches for image reconstruction.}}
\label{fig4}
\end{figure}
 As shown in Fig.~\ref{fig4}, we compare the reconstructed images with 75\% mask rate after 150 epoch training. Obviously, the contour of the reconstructed images by the contour-guided mask is more clear than the ones of MAE.
\begin{table}[!t]\small
\centering
\caption{Performance comparison of MAE and ConMAE on VeRi-776 dataset.}
\label{tab3}
\begin{tabular}{c|cccc}
\toprule[0.8pt]
method             & mAP        & Rank-1    & Rank-5     & Rank-10    \\ \hline
MAE        & 41.8          & 80.5          & 85.3          & 86.8          \\ 
ConMAE & \textbf{44.2} & \textbf{82.7} & \textbf{87.9} & \textbf{90.6} \\ \hline
\end{tabular}
\end{table}

The experimental results show that the mAP of the contour-guided mask is 2.4\% higher than that of the random mask, which proves the effectiveness of the contour-guided mask. In the task of vehicle re-identification, the vehicle contour is often the key region. The model performance is improved because it retains the vehicle contour effectively.

\subsubsection{Effect of Different Mask Rate}
In order to study the impact of different mask rates on the model performance, we carried out the experiments.

\begin{figure}[!t]
\centering
\includegraphics[width=\hsize]{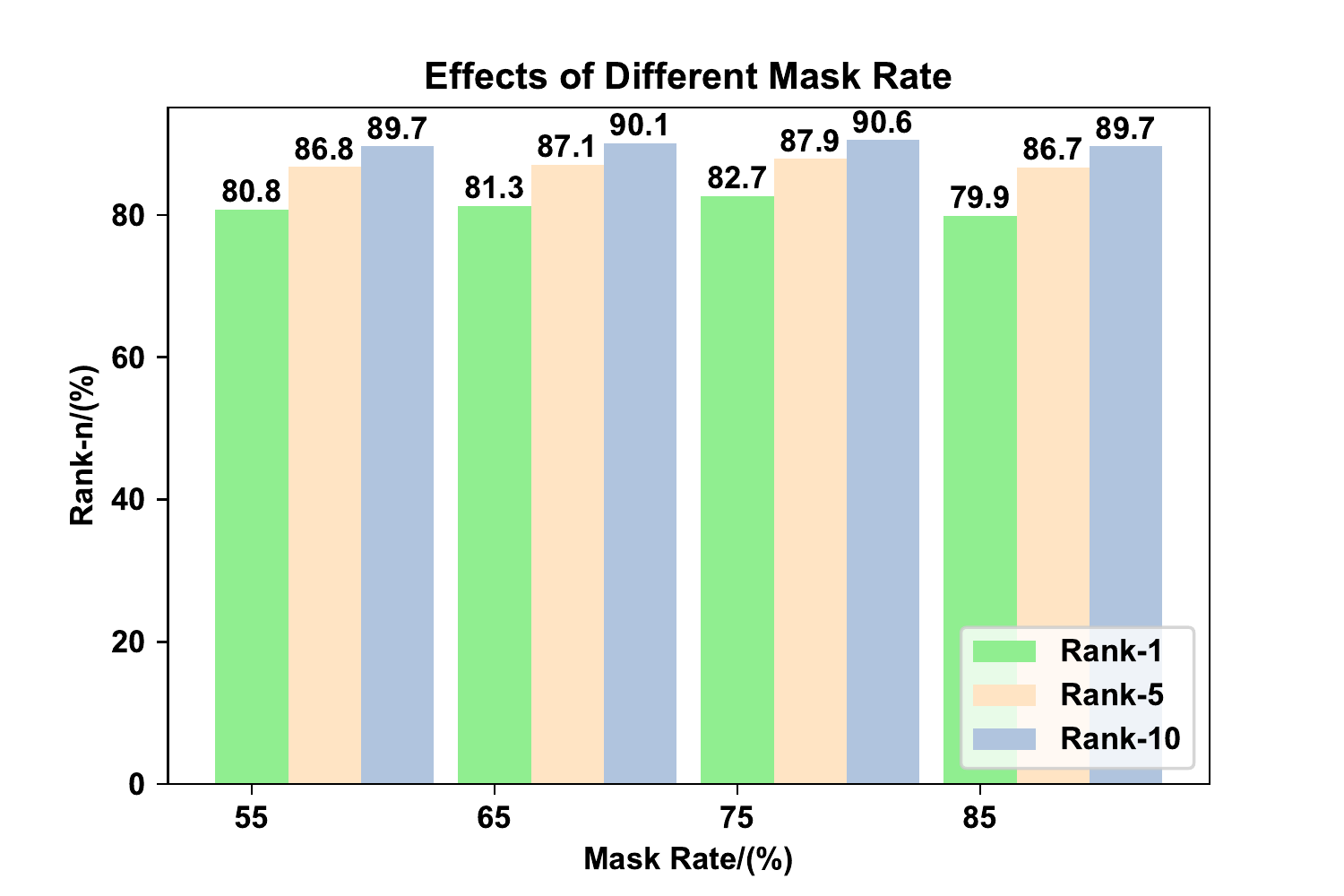}
\caption{Performance of different mask rates on the VeRi-776 dataset.}
\label{fig7}
\end{figure}

As shown in Fig.~\ref{fig7}, the experimental results show that the model performs best when the mask rate is 75\%. Mask rate has a significant impact on model performance. High mask rates, such as 85\%, make it difficult for ConMAE to reconstruct images, resulting in low feature extraction capability of pre-training models. A low mask rate, such as 55\%, on the one hand, causes too many image blocks in the input encoder, which requires additional computing costs, and on the other hand, leads to a decline in model accuracy.

\subsubsection{Effect of Progressive Softening Mechanism}
In order to study the influence of the progressive softening mechanism on the model accuracy, we carry out experiments for the soft or hard pseudo labels, and the results are shown in Table~\ref{tab4}.

\begin{table}[htpb]\small
\centering
\caption{Effects of soft and hard pseudo labels on VeRi-776 dataset.}
\label{tab4}
\begin{tabular}{c|cccc}
\toprule[0.8pt]
method              & mAP& Rank-1 & Rank-5 & Rank-10\\ \hline
hard label &35.6     &79.1        &84.7       &87.5        \\ 
soft label &\textbf{44.2}     &\textbf{82.7}        &\textbf{87.9}       &\textbf{90.6}         \\ \hline
\end{tabular}
\end{table}

As shown in the experimental results, the performance of soft pseudo labels is better than that of hard pseudo labels in all indicators. The hard pseudo labels generated by clustering will inevitably have errors, and the use of hard pseudo labels can minimize the impact of errors. Therefore, soft pseudo labels have better performance.

\section{Conclusion}

In this paper, MAE is applied to unsupervised vehicle re-identification, and the MAE of the random mask is transformed into contour-guided MAE. After self-supervised training, an encoder with good feature embedding ability is obtained. The encoder is applied to the unsupervised vehicle re-identification task so that the model can generate the pseudo labels with higher confidence. In addition, this paper also designs a progressive pseudo-label softening mechanism. The hard pseudo labels of the samples will be gradually optimized to soft pseudo labels in the training epoch, which reduces the impact of false pseudo labels on the accuracy of the model. The experimental results prove the effectiveness of contour-guided MAE for unsupervised vehicle re-identification.

\small{\bibliographystyle{IEEEtran}
\bibliography{ref}}
\end{document}